\documentclass[runningheads]{llncs}
\usepackage[T1]{fontenc}
\usepackage{graphicx,verbatim}
\usepackage{amsmath}
\usepackage{amssymb}
\usepackage{hyperref}
\usepackage{tabularx}
\usepackage{booktabs}
\usepackage{array}
\usepackage{placeins}
\usepackage{needspace}
\begin{document}
\title{Intervention-Aware Clinical World Model for Post--Op Outcome Forecasting in Cardiology}
\titlerunning{Intervention-Aware Clinical World Model}
%
\author{Yunsung Chung \inst{1} \and
Yingshuo Liu \inst{1} \and
Abboud F. Hassan \inst{1} \and
Han Feng \inst{1} \and
Mary M. Maleckar \inst{1,2} \and
Nassir Marrouche \inst{1} \and
Jihun Hamm \inst{1}\thanks{Corresponding author.}}
\authorrunning{Y. Chung et al.}
%
\institute{Tulane University, New Orleans, LA 70118, USA \\
\email{\{ychung3,jhamm3\}@tulane.edu}
\and
Simula Research Laboratory, Oslo, Norway}


  
\maketitle              
\begin{abstract}
Many clinical prediction models treat post-intervention outcomes as a one-step mapping from baseline measurements to a future endpoint.
However, recovery after a procedure often unfolds as an irregular trajectory: clinical observations, medication changes, repeat interventions, and physiological measurements are recorded asynchronously and can change risk assessment over time.
We propose an \emph{intervention-aware clinical world model} that represents each patient with a structured latent state and evolves it through time-ordered post-intervention events.
The model first encodes baseline imaging into a 3D spatial latent state. It then updates this state using procedural context, static covariates, elapsed time, and peri-event physiological embeddings. Follow-up imaging provides training-only supervision through a latent forecasting objective.
We apply the framework to atrial fibrillation ablation. During the 90-day recovery window, irregular post-procedure records provide clinically meaningful evidence for long-term recurrence risk.
In repeated internal cross-validation on DECAAF-II, our model achieves AUROC $0.756$ and AUPRC $0.777$ for recurrence prediction. It also achieves a scar-extent MAE of $2.971$ percentage points without requiring follow-up MRI intensities at inference.
The learned state supports recurrence-risk queries at different horizons and retrospective input editing of blanking-period records.

\end{abstract}

\section{Introduction}
Many medical imaging models map a baseline scan directly to a future endpoint~\cite{chung2025sofa}. Yet recovery is dynamic: medication changes and follow-up interventions occur at irregular times and can change risk estimates. Models should therefore update predictions as longitudinal evidence accrues.

\begin{figure}[t]
  \centering
  \includegraphics[width=0.9\textwidth]{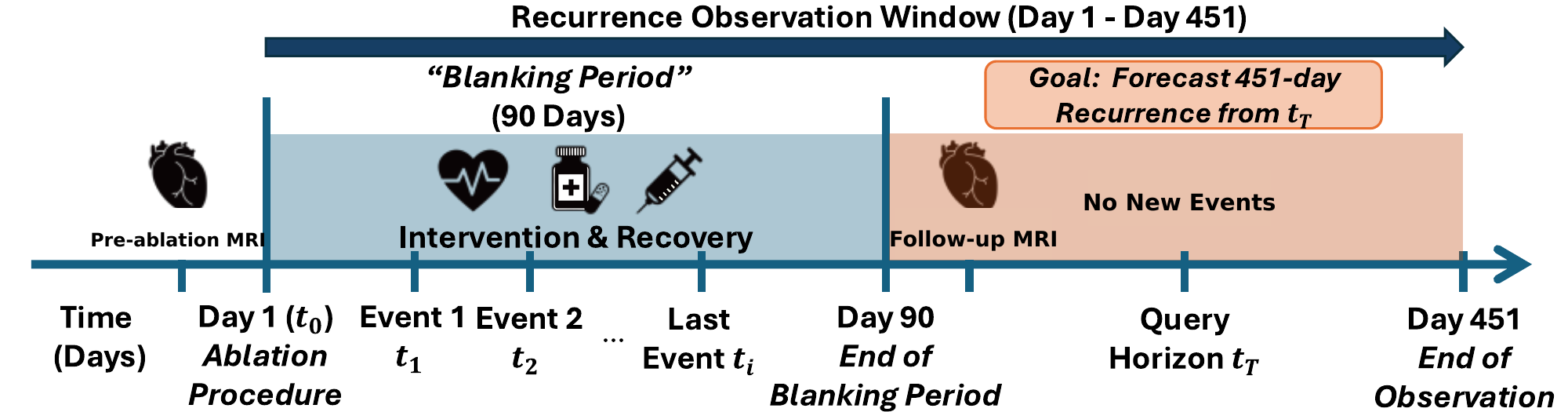}
  \caption{
    \textbf{Clinical timeline:} Irregular interventions during the 90-day blanking period necessitate dynamic risk updating rather than a single-shot baseline prediction.
    } 
  \label{fig:timeline}
\end{figure}

We study atrial fibrillation (AF) ablation. Pre-ablation LGE-MRI captures atrial fibrosis and anatomy. During the 90-day clinical ``blanking period,'' medication changes, electrical cardioversion, and occasional repeat procedures are recorded at irregular times~\cite{calkins20182017}. These events are rarely modeled with baseline imaging. We forecast recurrence within 451 days using only records available by a selected query horizon (Fig.~\ref{fig:timeline}).

To address this gap, we introduce an \emph{intervention-aware clinical world model}. It represents atrial anatomy in a spatial latent state, then updates the state with procedural geometry, static covariates, elapsed time, and pre-event ECG embeddings as blanking-period events occur. Rather than returning one fixed score, it provides retrospective risk estimates at selected horizons.

\begin{figure}[t]
  \centering
  \includegraphics[width=0.9\textwidth]{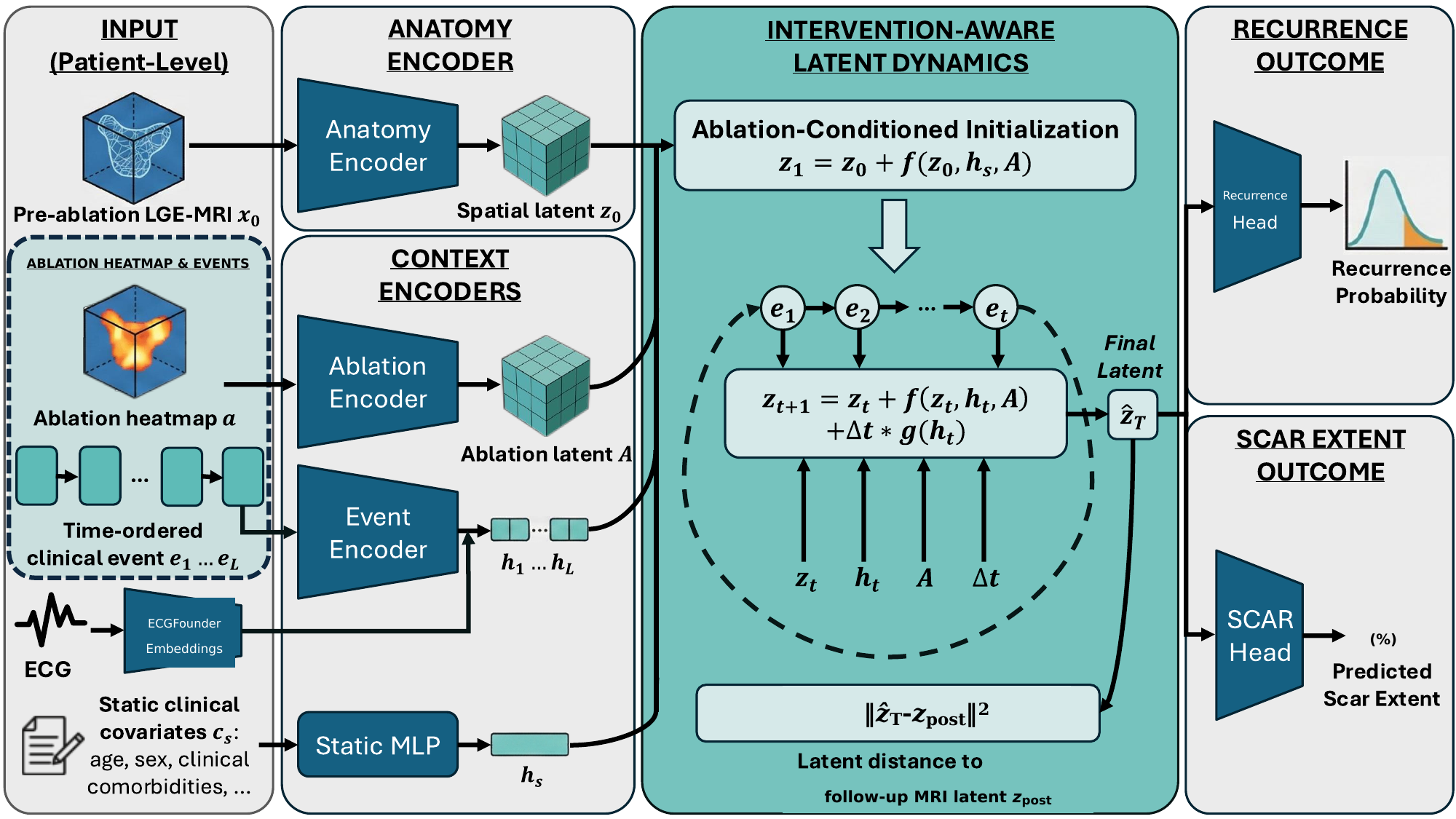}
  \caption{
    \textbf{Overview of the proposed intervention-aware clinical world model.}
    Pre-ablation LGE-MRI is encoded into a spatial latent state $z_0$.
    Static covariates, the ablation heatmap, and time-ordered clinical events with ECGFounder embeddings condition ablation-conditioned initialization, latent updates, and elapsed-time drift.
    The final latent $\hat z_T$ is matched to the training-only follow-up latent $z_{\mathrm{post}}=E_\theta(x_{\mathrm{post}})$ and predicts recurrence probability and scar extent.
    }
  \label{fig:pipeline}
\end{figure}

\noindent\textbf{Contributions.}
We make three contributions:
\begin{enumerate}
    \item An intervention-aware latent clinical world model that evolves a 3D anatomical state conditioned on ablation geometry, irregular post-procedural events, and ECG embeddings.
    \item A horizon-token formulation for anytime recurrence forecasting from partial blanking-period histories.
    \item Internal evaluation on the DECAAF-II complete-record cohort and a larger auxiliary cohort without usable ablation geometry, reported separately from the full-model evaluation.
\end{enumerate}

\section{Related Work}
\subsubsection{World models and medical world models.}
World models learn state transitions for forecasting, simulation, and planning~\cite{bruce2024genie,hafner2025mastering}. Predictive representation learning emphasizes compact latent dynamics that avoid brittle voxel-level reconstruction~\cite{bardes2024revisiting,burchi2024mudreamer}.
In medicine, generative modeling has increasingly shifted from visual realism toward clinically useful synthesis, translation, and task-aligned generation~\cite{zhou2025generative,chung2026craft}.
Recent medical world models include treatment-conditioned rollouts~\cite{qazi2025beyond}, diffusion-based post-treatment synthesis~\cite{yang2025medical}, context-aware latent transitions~\cite{ding2025clarity}, and long-horizon electronic health record (EHR) patient simulation~\cite{mu2026ehrworld}.
Related latent-transition ideas also appear in interactive imaging and radiology-focused predictive representation learning~\cite{jiang2024cardiac,yue2025echoworld,yue2025chexworld}.
In contrast, we focus on post-ablation structural trajectory forecasting by coupling 3D LGE-MRI, procedural geometry, irregular blanking-period events, and peri-event ECG embeddings.

\subsubsection{Atrial LGE-MRI and ablation outcome modeling.}
Prior AF ablation studies segment and quantify the left atrium (LA) and scar~\cite{li2022atrialjsqnet,lefebvre2022lassnet}. Other work combines imaging and clinical features to predict recurrence~\cite{varela2017novel,atta2021new,roney2022predicting,razeghi2023atrial}. Other studies~\cite{muffoletto2021toward,muizniece2021reinforcement,ogbomo2023exploring} examine patient-specific and reinforcement-learning approaches to ablation strategy.
SOFA~\cite{chung2025sofa} introduces action-conditioned scar simulation from ablation patterns.
Unlike static recurrence predictors or scar simulators, our work models the post-procedural blanking period as an irregular event sequence and updates long-horizon recurrence risk by evolving a latent anatomical state.

\section{Method}
\label{sec:method}

\subsubsection{Problem setup and targets.}
The pre-ablation MRI $x_0$ initializes the state; the follow-up MRI $x_{\mathrm{post}}$ is used only in training to define $z_{\mathrm{post}}=E_\theta(x_{\mathrm{post}})$.
A terminal horizon token sets the query time $t_T$; $T$ is the forecast horizon, not the follow-up scan time.
We forecast 451-day AF recurrence $y$ and scar extent $b$ using observations available by that horizon. Scar extent is the percentage of scar-positive voxels in the follow-up LA wall mask; MAE is reported in percentage points. Figs.~\ref{fig:timeline} and~\ref{fig:pipeline} summarize the timeline and model.

\subsubsection{Latent anatomical state from LGE-MRI.}
A frozen variational autoencoder (VAE) $(E_\theta, D_\theta)$, pretrained on a larger imaging cohort, maps each MRI to a 3D spatial latent state
$z=E_\theta(x)\in\mathbb{R}^{C\times d\times h\times w}$.
We set the initial state to $z_0=E_\theta(x_0)$ and the training-only follow-up target to $z_{\mathrm{post}}=E_\theta(x_{\mathrm{post}})$.

\subsubsection{Conditioning signals and event tokens.}
Static covariates are embedded as $h_s=\phi_s(c_s)$, and a 3D convolutional neural network (CNN) encodes the ablation heatmap as
$A=\phi_a(a)\in\mathbb{R}^{C_a\times d\times h\times w}$.
Each event at time $t_i$ is mapped to a token 
\begin{equation}
e_i=\Big[
\tfrac{t_i-t_{\mathrm{abl}}}{90},\;
\tfrac{t_T-t_{\mathrm{abl}}}{180},\;
b_{\mathrm{med}},\,b_{\mathrm{cv}},\,b_{\mathrm{rep}},\;
\psi_{\mathrm{med}}(\mathrm{med_{cat}}),\;
u_i
\Big],
\end{equation}
The first two scalar entries encode normalized event time and normalized queried horizon, respectively.
The three binary flags indicate medication, cardioversion, or a repeat procedure. The function $\psi_{\mathrm{med}}$ embeds either a medication category or a dedicated \texttt{none} token.
ECG context $u_i$ mean-pools ECGFounder~\cite{li2024electrocardiogram} embeddings recorded within 7 days before $t_i$. This window balances temporal relevance against sparse acquisition. Using only pre-event recordings avoids future information, and mean pooling handles variable recording counts. We set $u_i=\mathbf{0}$ when no ECG is available.

We prepend an ablation anchor token at $t_{\mathrm{abl}}$, sort clinical events by recorded time, censor them at day 90, and append the terminal horizon token at $t_T$ with zero event flags. Variable-length sequences are padded and masked by $m_i\in\{0,1\}$.
A Transformer encoder produces contextualized embeddings
\begin{equation}
h_{1:L_{\max}}=E_c(e_{1:L_{\max}},m_{1:L_{\max}}),
\end{equation}
For each token, we form $c_i=\phi_c([h_s\|h_i])$. These conditioning vectors are broadcast over the latent grid for the 3D convolutional transitions.

\subsubsection{Event-conditioned latent dynamics.}
The first update ($i{=}1$) uses the ablation anchor and encoded ablation map $A$. Later updates incorporate event context and elapsed-time drift.
Starting from $\hat z_0=z_0$, we update the latent state once per valid token $i$:
\begin{align}
\Delta t_i &= \frac{\max(t_i-t_{i-1},0)}{\tau}, \quad t_0=t_{\mathrm{abl}},\\
\hat z_i &= \hat z_{(i-1)} + f_\phi(\hat z_{(i-1)},c_i,A) + \Delta t_i\,g_\phi(c_i),
\label{eq:update_compact}
\end{align}
Here, $f_\phi$ is a 3D residual CNN, while the multilayer perceptron (MLP) $g_\phi$ produces a context-dependent drift broadcast over space. They do not share parameters, and $\tau{=}30$ days.
The terminal token advances the state without adding a clinical event.
To query a new horizon, we change only the time in this terminal token. Observed time gaps represent irregular sampling. Missing ECG is zero-filled, padding is masked, and unrecorded events are not imputed.

\subsubsection{Outcome heads and training objective.}
We spatially average the predicted terminal latent $\hat z_T$ and encoded ablation map $A$ to obtain vector summaries $\bar z$ and $\bar a$. These summaries are concatenated with static context for the recurrence and scar heads.
We train $(f_\phi,g_\phi)$ and the heads with the loss
\begin{equation}
\mathcal{L}=
\lambda_z\|\hat z_T-z_{\mathrm{post}}\|_2^2
+\lambda_{\mathrm{cls}}\mathcal{L}_{\mathrm{BCE}}(\hat p,y)
+\lambda_{\mathrm{bur}}\mathcal{L}_{\mathrm{Huber}}(\hat b,b),
\label{eq:loss_compact}
\end{equation}
where $\mathcal{L}_{\mathrm{BCE}}$ is binary cross-entropy, $\mathcal{L}_{\mathrm{Huber}}$ is Huber loss, and $z_{\mathrm{post}}$ encodes the follow-up MRI. Inference excludes follow-up MRI; inputs are pre-ablation MRI, the ablation heatmap, static covariates, the event prefix, and available pre-event ECG embeddings. Later forecasts change only the terminal horizon token.

\begin{table}[t]
\centering
\scriptsize
\setlength{\tabcolsep}{3pt}
\renewcommand{\arraystretch}{1.12}
\caption{Cohort construction and available modalities.}
\label{tab:cohort_availability}
\begin{tabularx}{\linewidth}{@{}
>{\raggedright\arraybackslash}p{0.20\linewidth}
c
>{\raggedright\arraybackslash}p{0.20\linewidth}
>{\raggedright\arraybackslash}X
@{}}
\toprule
\textbf{Cohort} & \textbf{N} & \textbf{Use} & \textbf{Available data} \\
\midrule
Parent union
& 830
& Availability audit
& Patients linked across available sources \\

Complete full-modality cohort
& 91
& Full-model evaluation
& Pre/post LGE-MRI, ablation geometry, blanking-period events, ECGs, scar, recurrence \\

Auxiliary cohort without usable ablation geometry
& 258
& Evaluation without ablation map
& Pre-MRI, static covariates, events, ECGs, recurrence; no usable ablation map \\

\midrule
Complete-cohort events
& --
& Event schema
& Medication 117, cardioversion 51, repeat procedure 3; median 2 events/patient \\
\bottomrule
\end{tabularx}
\end{table}

\section{Experiments and Results}

\subsection{Datasets and Preprocessing}
\subsubsection{Dataset and evaluation.} 
We use the DECAAF-II~\cite{marrouche2022effect} cohort, a multicenter study with paired pre- and post-ablation LGE-MRI and procedural records that include ablation points.
We pretrain $E_\theta, D_\theta$ on a \emph{disjoint} set of $N{=}732$ patients with LGE-MRI but incomplete ablation or event records. The state-transition model is evaluated on $N{=}91$ complete-record patients using 5-fold cross-validation with 3 seeds. Table~\ref{tab:cohort_availability} summarizes data availability and event frequency.
For recurrence, we report the areas under the receiver operating characteristic and precision-recall curves (AUROC and AUPRC). For scar extent, we report MAE.

\subsubsection{Preprocessing.}
Electroanatomic mapping (EAM) ablation points are manually mapped to MRI space and converted to a 3D heatmap. We resample both LGE-MRIs to 1.5\,mm isotropic resolution, rigidly register the post-ablation scan to the pre-ablation scan, and transform the post-ablation masks accordingly. Volumes are cropped to the pre-ablation LA bounding box with a 10\,mm margin and center-padded or cropped to $(80,80,96)$. Intensities are normalized to $[-1,1]$ using robust region-of-interest (ROI) percentiles from the pre-ablation scan; the same parameters are applied to the registered post-ablation scan.
Additional preprocessing details are provided in the \href{https://github.com/cys1102/af-blanking-world-model}{code repository}.

\begin{table}[t]
\centering
\scriptsize
\setlength{\tabcolsep}{3.5pt}
\caption{Complete-record results ($N=91$; 5-fold CV, 3 seeds; mean$\pm$SD). Post-MRI-only uses follow-up MRI at inference; sequence baselines use our allowed inputs.}
\label{tab:main_results}
\begin{tabular}{lccc}
\toprule
\textbf{Method} & \textbf{AUROC} $\uparrow$ & \textbf{AUPRC} $\uparrow$ & \textbf{Scar MAE} $\downarrow$ \\
\midrule
Static-only & $0.511\pm0.126$ & $0.551\pm0.070$ & $6.507\pm1.667$ \\
Post-MRI-only & $0.525\pm0.045$ & $0.575\pm0.079$ & $3.189\pm0.698$ \\
ECGFounder w/ECG & $0.639\pm0.134$ & $0.635\pm0.133$ & $4.773\pm1.403$ \\
SOFA & $0.610\pm0.142$ & $0.660\pm0.134$ & $3.510\pm0.903$ \\
\midrule
  GRU & $0.611\pm0.157$ & $0.661\pm0.137$
  & $11.920\pm11.033$ \\
  LSTM & $0.653\pm0.132$ & $0.687\pm0.106$
  & $9.146\pm10.317$ \\
  Transformer & $0.596\pm0.142$ &
  $0.668\pm0.114$ & $9.171\pm9.280$ \\
\midrule
\textbf{Ours} & $\mathbf{0.756}\pm\mathbf{0.051}$ & $\mathbf{0.777}\pm\mathbf{0.058}$ & $\mathbf{2.971}\pm\mathbf{0.675}$ \\
\bottomrule
\end{tabular}
\end{table}

\begin{table}[t]
\centering
\scriptsize
\setlength{\tabcolsep}{3.5pt}
\caption{Auxiliary no-geometry results ($N=258$; 5-fold CV, 3 seeds; mean$\pm$SD). Sequence baselines use the same inputs; this is not full-model validation.}
\label{tab:auxiliary_results}
\begin{tabular}{lccc}
\toprule
\textbf{Method} & \textbf{Inputs} & \textbf{AUROC} $\uparrow$ & \textbf{AUPRC} $\uparrow$ \\
\midrule
Static-only & Static & $0.517\pm0.073$ & $0.571\pm0.074$ \\
Pre-MRI-only & MRI latent & $0.529\pm0.081$ & $0.594\pm0.078$ \\
ECGFounder-only & ECG & $0.596\pm0.039$ & $0.608\pm0.038$ \\
GRU & MRI+static+events+ECG & $0.555\pm0.059$ & $0.592\pm0.063$ \\
LSTM & MRI+static+events+ECG & $0.552\pm0.062$ & $0.584\pm0.071$ \\
Transformer & MRI+static+events+ECG & $0.549\pm0.072$ & $0.581\pm0.069$ \\
\midrule
\textbf{Ours w/o ablation map} & MRI+static+events+ECG & $\mathbf{0.713}\pm\mathbf{0.088}$ & $\mathbf{0.747}\pm\mathbf{0.078}$ \\
\bottomrule
\end{tabular}
\end{table}

\begin{table}[t]
\centering
\scriptsize
\caption{Component ablations on conditioning signals and model components.}
\label{tab:ablations}
\setlength{\tabcolsep}{5pt}
\begin{tabular}{lcc}
\hline
Method & AUROC $\uparrow$ & AUPRC $\uparrow$ \\
\hline
\textbf{Ours (full)} & $\mathbf{0.756}\pm\mathbf{0.051}$ & $\mathbf{0.777}\pm\mathbf{0.058}$  \\
\hline
\multicolumn{3}{l}{\textit{Conditioning ablations}} \\
w/o terminal token & $0.742\pm0.064$ & $0.738\pm0.070$ \\
w/o AF ablation map & $0.711\pm0.058$ & $0.758\pm0.086$ \\
w/o ECGFounder & $0.705\pm0.136$ & $0.748\pm0.131$ \\
w/o events & $0.623\pm0.121$ & $0.652\pm0.111$ \\
\hline
\multicolumn{3}{l}{\textit{Objective / architecture ablations}} \\
w/o latent matching loss ($\lambda_z{=}0$) & $0.624\pm0.200$ & $0.686\pm0.173$ \\
Direct fusion (no step-wise evolution) & $0.709\pm0.095$ & $0.720\pm0.077$ \\
\hline
\end{tabular}
\end{table}

\subsection{Results}
\begin{figure}[t]
\centering
\includegraphics[width=0.9\linewidth]{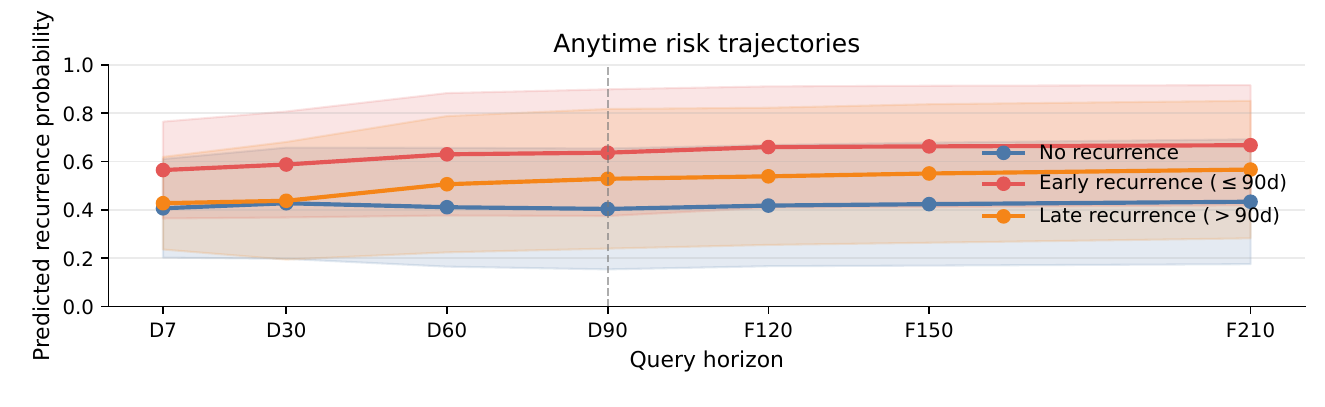}
\caption{\textbf{Anytime risk trajectories.}
Mean$\pm$SD for early recurrence ($\le$90d, $n{=}32$), late recurrence ($>$90d, $n{=}14$), and no recurrence ($n{=}45$). D: observed-day; F: no-new-event horizons.}
\label{fig:temporal_prediction}
\end{figure}

\subsubsection{Forecasting.}
Using blanking-period information, we forecast diagnosis within 451 days post-ablation. In Table~\ref{tab:main_results}, Static-only and Post-MRI-only remain near chance, while ECGFounder and LSTM improve discrimination. Our $0.756/0.777$ AUROC/AUPRC improves on matched-input LSTM by $0.103/0.090$, indicating that state evolution, not input access alone, matters. OOF Brier score $0.201$ and five-bin expected calibration error $0.032$ show close coarse-bin probability alignment internally.

Fig.~\ref{fig:temporal_prediction} shows that separation develops as blanking-period evidence accumulates: from D30 to D90, risk rises fastest for early recurrence, remains comparatively stable for no recurrence, and is intermediate for late recurrence. The ordering changes only gradually during F120--F210 no-new-event rollouts, indicating that the observed record establishes most of the separation. Overlapping SD bands, especially for late recurrence, nevertheless show substantial patient-level overlap.

\Needspace{7\baselineskip}
Scar extent provides a structural check on the latent trajectory. Our method reaches MAE $2.971\pm0.675$ percentage points versus $3.189\pm0.698$ for oracle-style Post-MRI-only, despite not receiving follow-up MRI intensities at inference. The $0.218$-point margin is small relative to fold variability, so the finding is competitive structural forecasting rather than reliable superiority over the oracle-style baseline. Relative to the cohort mean (interquartile range) of $8.61\%$ ($5.75$--$10.69\%$), the error remains nontrivial; no inter-observer benchmark is available.

Because complete multimodal records with usable ablation geometry are scarce, we use $N=258$ additional patients to test whether the common-input rollout signal extends beyond the $N=91$ cohort. Without an ablation map, our variant obtains AUROC/AUPRC $0.713/0.747$, gains of $0.158/0.155$ over the strongest matched-input sequence baseline (Table~\ref{tab:auxiliary_results}). These values resemble the $N=91$ no-map ablation ($0.711/0.758$), although the cohorts are not directly comparable. The consistency indicates that event-conditioned rollout retains signal without geometry, while leaving the full model's added spatial value untested in the larger cohort.

\subsubsection{Ablations.}
Table~\ref{tab:ablations} reveals a contribution hierarchy: removing events or latent matching produces the largest AUROC reductions ($0.133$ and $0.132$), while removing the ablation map or ECG has smaller effects. Without latent matching, AUROC SD also rises from $0.051$ to $0.200$, consistent with structural supervision stabilizing the learned state. Direct fusion loses $0.047$ AUROC and $0.057$ AUPRC, supporting step-wise evolution over one-shot concatenation.

Shuffling event content while preserving times reduces AUROC/AUPRC from $0.756/0.777$ to $0.709/0.713$. Because event times and counts remain fixed, the drop indicates that event identity contributes beyond elapsed time.

\begin{figure}[t]
\centering
\includegraphics[width=0.9\linewidth]{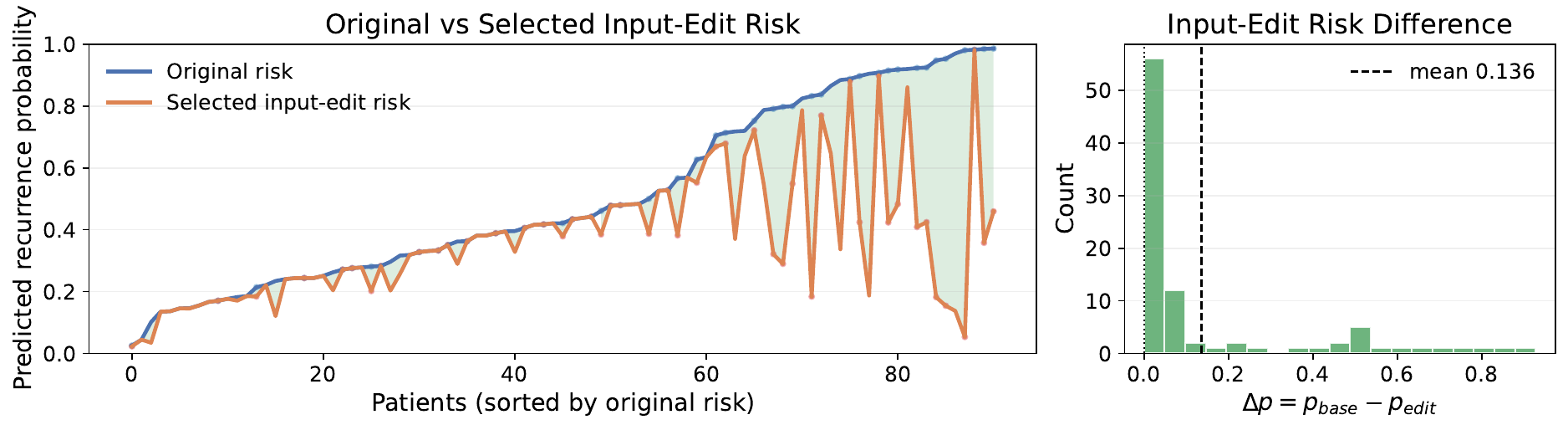}
\caption{\textbf{Input-editing sensitivity.}
\textbf{Left:} patient-wise $\Delta p=p_{\mathrm{base}}-p_{\mathrm{edit}}$ after predefined edits.
\textbf{Right:} cohort distribution of $\Delta p$.
Edits are associational probes, not causal treatment effects.}
\label{fig:input_edit_risk_trajectory}
\end{figure}

\Needspace{7\baselineskip}
Fig.~\ref{fig:input_edit_risk_trajectory} gives a second view of how the model uses blanking-period evidence.
Specifically, we edit the presence or timing of medication, cardioversion, and repeat-procedure events and recompute recurrence risk, testing the model's ability to compare alternative blanking-period scenarios.
Edited and observed risks remain close for most patients, while a smaller subset with higher baseline risk shows large downward shifts. The resulting right-skewed $\Delta p$ distribution has mean $0.136$, but this value is subset-driven rather than a uniform cohort shift. The fitted model's input dependence is therefore heterogeneous and not captured by the mean alone.

\FloatBarrier

\subsection{Limitations}
The complete-record cohort is small ($N{=}91$), and all results are internal to DECAAF-II. The $N{=}258$ cohort lacks ablation geometry and cannot replace external validation. Manual EAM-to-MRI mapping can introduce localization error, while EAM measurements themselves may be noisy. Because clinical events may be missing or confounded, input edits show model sensitivity rather than causal effects. Alternative ECG windows, the effect of missingness patterns, latent-loss effects on scar MAE, and inter-observer scar variability remain unevaluated.

\section{Conclusion}
We presented an intervention-aware clinical world model that evolves a structured 3D latent state from ablation geometry, irregular blanking-period events, and ECG embeddings. In internal DECAAF-II cross-validation, it improves recurrence discrimination over static and matched-input sequence baselines, predicts scar extent without follow-up MRI at inference, and supports multi-horizon risk updates and retrospective input edits. Larger full-modality cohorts and external validation are needed before extending the framework toward causal modeling of post-ablation interventions.

    



%
%
%
\bibliographystyle{splncs04}
\bibliography{miccai2026}
%




\end{document}